\documentclass{sig-alternate-05-2015}

\usepackage{booktabs}
\usepackage{multirow}
\usepackage{graphics}
\usepackage{graphicx}
\usepackage{tabularx}
\usepackage{times}
\usepackage{url}
\usepackage{color}

\usepackage{enumitem}

\newenvironment{packed-enum}{
\begin{itemize}
  \setlength{\itemsep}{1pt}
    \setlength{\parskip}{0pt}
      \setlength{\parsep}{0pt}
}{\end{itemize}}

\newenvironment{packed-enum3}{
\begin{itemize}[leftmargin=*]
  \setlength{\itemsep}{1pt}
    \setlength{\parskip}{0pt}
      \setlength{\parsep}{0pt}
}{\end{itemize}}

\newenvironment{packed-desc}{
\begin{description}
  \setlength{\itemsep}{1pt}
    \setlength{\parskip}{0pt}
      \setlength{\parsep}{0pt}
}{\end{description}}

\newenvironment{packed-enum2}{
\begin{enumerate}[leftmargin=*]
  \setlength{\itemsep}{.5pt}
    \setlength{\parskip}{0pt}
      \setlength{\parsep}{0pt}
}{\end{enumerate}}

\begin{document}

\setcopyright{acmcopyright}



\conferenceinfo{Communications of the ACM}{In Press}


%

\title{Healthcare Robotics}

%
%
%
%
%

\numberofauthors{1} 
%
\author{
%
%
\alignauthor
Laurel D. Riek \\
       \affaddr{Computer Science and Engineering}\\
       \affaddr{University of California, San Diego}\\
       \email{lriek@ucsd.edu}
}


\maketitle
\begin{abstract}
Robots have the potential to be a game changer in healthcare: improving health and well-being, filling care gaps, supporting care givers, and aiding health care workers. However, before robots are able to be widely deployed, it is crucial that both the research and industrial communities work together to establish a strong evidence-base for healthcare robotics, and surmount likely adoption barriers. This article presents a broad contextualization of robots in healthcare by identifying key stakeholders, care settings, and tasks; reviewing recent advances in healthcare robotics; and outlining major challenges and opportunities to their adoption.
\end{abstract}

%
%


%
%

%
%



\section{Introduction}

The use of robots in healthcare represents an exciting opportunity to help a large number of people.
Robots can be used to enable people with cognitive, sensory, and motor impairments, help people who are ill or injured, support caregivers, and aid the clinical workforce. This article highlights several recent advancements on these fronts, and discusses their impact on stakeholders. It also outlines several key technological, logistical, and design challenges faced in healthcare robot adoption, and suggests possible avenues for overcoming them.


Robots are ``physically embodied systems capable of enacting physical change in the world``. They enact this change with effectors which can move the robot  (locomotion), or objects in the environment (manipulation). Robots typically use sensor data to make decisions. They can vary in their degree of autonomy, from fully autonomous to fully teleoperated, though most modern systems have mixed initiative, or shared autonomy. More broadly, \textit{robotics technology} includes affiliated systems, such as related sensors, algorithms for processing data, etc. \cite{riek-mental-health}.

There have been many recent exciting examples of robotics technology, such as autonomous vehicles, package delivery drones, and robots that work side-by-side with skilled human workers in factories. 
One of the most exciting areas where robotics has a tremendous potential to make an impact in our daily lives is in healthcare. 

An estimated 20\% of the world's population experience difficulties with physical, cognitive, or sensory functioning, mental health, or behavioral health. These experiences may be temporary or permanent, acute or chronic, and may change throughout one's lifespan. Of these individuals, 190 million experience severe difficulties with activities of daily living tasks (ADL) \footnote{World Bank, 2011 \url{http://documents.worldbank.org/ curated/en/2011/01/14440066/world-report-disability}}. These include physical tasks (basic ADLs), such as grooming, feeding, and mobility, to cognitive functioning tasks (instrumental ADLs), which include goal-directed tasks such as problem solving, finance management, and housekeeping \cite{graf2008lawton}.
The world also has a rapidly aging population, who will only add to this large number of people who may need ADL help. Of all of these individuals, few want to live in a long-term care facility. Instead, many people would prefer to live and age gracefully in their homes for as long as possible, independently, and with dignity \cite{milligan2012there}.  However, for people requiring help with ADL tasks, this goal is challenging to meet for a few reasons. First, this level of care is quite expensive; in the US it costs between \$30,000 and \$85,000 per year in provider wages alone\footnote{U.S. Department of Health and Human Services, \url{http://longtermcare.gov/costs-how-to-pay/costs-of-care/}}.

\begin{figure*}[t!]
\centering
\includegraphics[width=\textwidth]{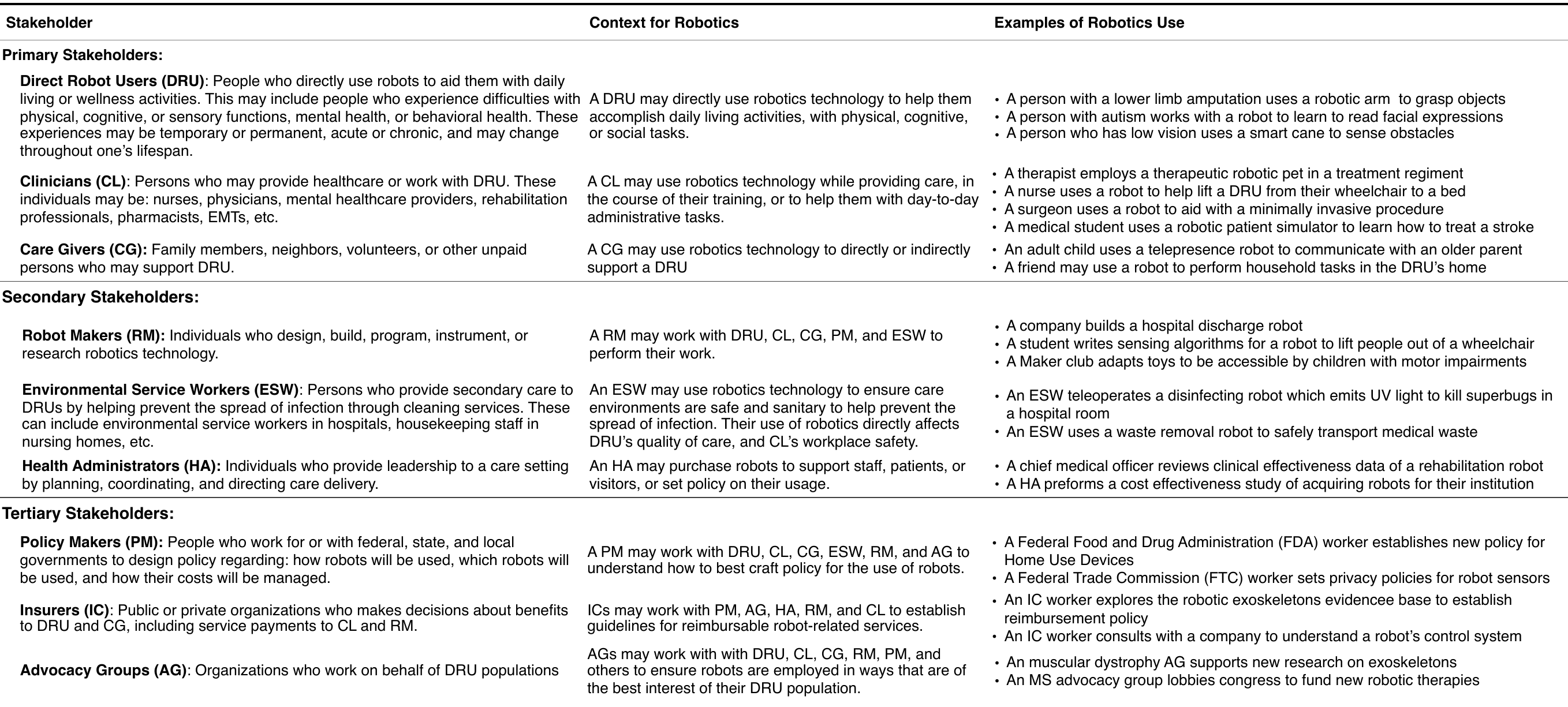}
\caption{The main  stakeholders for healthcare robotics, and exemplar contextualizations of their relationship to the technology.}
\label{tab:stakeholders}
\end{figure*}

Second, there is a substantial healthcare labor shortage -- there are far more people who need care than health care workers available to provide it  \cite{shi2014delivering}. While family members and friends attempt to fill these care gaps, they too have full-time jobs and other familial obligations, and thus cannot meet the need. 
Health care workers are not only overburdened by this labor shortage, but face increasingly hazardous work environments, and are themselves at great risk of debilitating injury and disability.  According to the National Institute for Occupational Health and Safety (NIOSH), health care workers have the most hazardous industrial jobs in America, with the greatest number of nonfatal occupational injuries and illness\footnote{National Institute for Occupational Safety and Health, \url{http:// www.cdc.gov/ niosh/ topics/ healthcare/}}.

Thus, there is an incredible opportunity for robotics technology to help fill care gaps and help aid health care workers. In both the research and commercial space, robotics technology has been used for physical and cognitive rehabilitation, surgery, telemedicine, drug delivery, and patient management. Robots have been used across a range of environments, including hospitals, clinics, homes, schools, and nursing homes; and in both urban and rural areas. 

Before discussing these applications, it is important to first contextualize the use of robots within healthcare. Section \ref{stake} identifies who will be providing, receiving, and supporting care, where this care will take place, and key tasks for robots within these settings. Then, Section \ref{section3} introduces examples of new technologies aimed at supporting these stakeholders, and the potential to meet their needs. 

Section \ref{adopt} outlines key challenges and opportunities to realizing the potential use of robots in healthcare, which the research and industrial communities are encouraged to consider moving forward. These adoption issues include consideration of a robot's capability and function (does a robot have the required capabilities to perform its function?), cost effectiveness (what is the robot's value to stakeholders relative to its cost?), clinical effectiveness (has the robot been shown to have a benefit to stakeholders?), usability and acceptability (How easy is the robot to use, modify, and maintain? Is the robot's form and function acceptable?), and safety and reliability (how safe and reliable is the robot?).

\section{Stakeholders, Care Settings, \\and Robot Tasks}
\label{stake}
\subsection{Stakeholders}

For this article, stakeholders are defined as people who have a vested interest in the use of robotics technology in healthcare. Stakeholders can be: people who directly use robots to provide assistance with daily living or wellness activities (direct robot users (DRU)), health professionals who use robots to provide care (clinicians (CL)), non-CL individuals who support DRUs (care givers (CG)), technologists and researchers (robot makers (RM)), health administrators (HAs), policymakers (PMs), advocacy groups (AGs), and insurers (IC). Fig. \ref{tab:stakeholders} introduces these stakeholders. 

These stakeholders can be grouped into three beneficiary groups.
1) Primary beneficiaries: direct robot users, clinicians, and caregivers, all of whom are likely to  use robotics technology on a regular basis; 2) Secondary beneficiaries: health administrators, robot makers, and environmental service workers, all of whom are involved in the use of robotics technology in healthcare settings but do not directly use the robots to use robots to support the health and wellness of DRUs; 3) tertiary beneficiaries: policy makers and advocacy groups, who have interest in the use of robots to provide care to their constituents, but are unlikely to use them directly. 

This article will focus on primary beneficiaries; however, it is important to note that all other stakeholder groups are critical to the successful end-deployment of robotics in health care, and should be included when possible in decision making.

\subsection{Care Settings}
\label{place}

Another critical dimension to contextualizing the use of robotics in healthcare is to consider the location of use. This can significantly impact on how suitable different technologies are for a given setting \cite{gonzales2015designing}, and can affect the design of a robot and its required capabilities. For example, while a 400lb, 5'4'' dual-arm mobile manipulator may work well in a lab, it is ill-suited to an 80 sq. ft. room in an assisted living facility.  While it is understandable robot makers may immediately be more concerned with achieving platform functionality than the particulars of care settings, to successfully deploy healthcare robots, setting must be considered. 

Table \ref{table:setting} defines different kinds of care settings, and includes longer-term care facilities in the community, as well as shorter-term care facilities, such as hospitals. For longer-term care in the US, the Fair Housing Act and Americans with Disabilities Act set some general guidelines for living space accessibility; however, the majority of space guidelines are state-dependent, and can have a large degree of variation. For example, an assisted living facility in Florida must provide 35 sq. ft. per resident for living and dining, whereas in Utah it is 100 sq. ft.. An in-patient psychiatric facility in Kentucky must provide 30 sq. ft. per patient in social common areas, Oregon requires 120 sq. ft. in total and 40 sq. ft. per patient.


Robots in healthcare can also affect the well-being, health, and safety of both direct robot users and clinicians.  The field of evidence-based healthcare design \cite{ulrich08} has produced hundreds of studies showing a relationship between the built environment and health and wellness, in areas including patient safety, patient outcomes, and staff outcomes. When new technology such as a robot becomes part of a care setting, it is now a possible disruptor to health. HAs must balance the risks and benefits for adopting new technology, and robot makers should be aware of these tradeoffs in how they design and test their systems.

{\subsection{Care Tasks}
\label{tasks}

Robots may be helpful for many health tasks. Robots can provide both physical and cognitive task support for both DRUs and clinicians/care givers, and may be effective and helping reduce cognitive load. Task assistance is particularly critical as the demand for healthcare services is far outpacing available resources, which places great strain on clinicians and care givers \cite{shi2014delivering}.}

\begin{table}[t!]
\centering
\begin{tabularx}{\columnwidth}{|X|}
\toprule
       
\textbf{Longer term care settings:}

\begin{packed-enum3}
\item \textbf{Assisted Living Facility}: ``Congregate residential facility with self-contained living units providing assessment of each resident's needs and on-site support 24 hours a day, 7 days a week, with the capacity to deliver or arrange for services including some health care and other services.''


\item \textbf{Group Home}: ``A residence, with shared living areas, where clients receive supervision and other services such as social and/or behavioral services, custodial service, and minimal services (e.g., medication administration).''


\item \textbf{Custodial Care Facility}: ``provides room, board and other personal assistance services, generally on a long-term basis, and which does not include a medical component.''


\item \textbf{Nursing Facility}: ``primarily provides to residents skilled nursing care and related services for the rehabilitation of injured, disabled, or sick persons, or, on a regular basis, health-related care services above the level of custodial care to other than [people with intellectual disabilities].''


\item \textbf{Home Care}: ``Location, other than a hospital or other facility, where [a person] receives care in a private residence.''
\end{packed-enum3}

\textbf{Shorter term care settings:}

\begin{packed-enum3}
\item \textbf{Inpatient Hospital}: ``A [non-psychiatric] facility, which primarily provides diagnostic, therapeutic (both surgical and nonsurgical), and rehabilitation services by, or under, the supervision of physicians to patients admitted for a variety of medical conditions.''


\item \textbf{On/Off Campus Outpatient Hospital}: ``A portion of a... hospital provider based department which provides diagnostic, therapeutic (both surgical and nonsurgical), and rehabilitation services to sick or injured persons who do not require hospitalization or institutionalization.''


\item \textbf{Inpatient Psychiatric Facility}: provides 24-hour ``inpatient psychiatric services for the diagnosis and treatment of mental [health disorders], by or under the supervision of a physician.''


\item \textbf{Substance Abuse Treatment facility}: ``provides treatment for substance (alcohol and drug) abuse on an ambulatory basis. [Provides] individual and group therapy and counseling, family counseling, laboratory tests, drugs and supplies, and psychological testing.'' Residential facilities also provide room and board.


\item \textbf{Hospice}: ``A facility, other than a patient's home, in which palliative and supportive care for terminally ill patients and their families are provided.''
\end{packed-enum3}
\\\toprule

\end{tabularx}

\caption{Selected care settings where robots may be used. Source: \textit{\protect\url{http://www.cms.gov/Medicare/Coding/place-of-service-codes/Place_of_Service_Code_Set.html}}
}
\label{table:setting}
\end{table}

\subsection{Physical Tasks}
\subsubsection{Clinicians}

Tasks involving the ``3Ds'' of robotics -- dirty, dangerous, and dull -- can be of particular value for clinical staff. Clinicians spend an inordinate amount of time on ``non-value added'' tasks, e.g., time away from treating patients. The overburden of these tasks creates a climate for error; so robots which can help clinicians effectively surmount these challenges would be a boon.  Some of these non-value added tasks include: 
\textit{Transportation}, such as moving materials or people from one place to another, 
\textit{Inventory}, such as patients waiting to be discharged, 
\textit{Search Time}, such as looking for equipment or paperwork,
\textit{Waiting}, for patients, materials, staff, medications, and
\textit{Overburdening of Staff and Equipment}, such as during peak surge times in hospitals  
\cite{wellman2016leading}.

Two of the best tasks for robots in this task space is material transportation and scheduling, which robots can be exceptionally skilled at given the right parameters. For example, robots that can fetch supplies, remove waste, and clean rooms. Another task robots can do that will help greatly improve the workplace for clinicians is moving patients. This is a very hazardous task - hospital workers, home health workers, and ambulance workers experience musculoskeletal injuries between three and five times the national average when moving patients according to NIOSH.

Robots can also help clinicians with other dangerous tasks, such as helping treat patients with highly infectious diseases. Robot-mediated treatment has become particularly pertinent after the recent Ebola outbreak, where clinicians and care givers can perform treatment tasks via telepresence robots \cite{kraft2016}. 

Finally, robots may help extend the physical capabilities of clinicians. For example, in surgical procedures, robots may provide clinicians with the ability to perform less invasive procedures to areas of the body inaccessible with existing instrumentation due to tissue or distance constraints. These can include types of neurological, gastric, and fetal surgical procedures \cite{webster2006toward}.

\subsubsection{Direct Robot Users}

When designing robots for DRUs, there is great value in designing straightforward solutions to problems. At a recent workshop discussing healthcare robotics, people with Amyotrophic Lateral Sclerosis (ALS) and other conditions reported that most of all they just wanted ``a robot to change the oil'' \cite{hri-ethics-workshop}. In other words: help is most needed with basic, physical ADL tasks, such as dressing, eating, ambulating, toileting, and housework. Robots that can help people avoid falling could also be incredibly beneficial, as falls cause thousands of fatal and debilitating injuries per year.

Currently, standalone robots that can successfully perform the majority of these key physical ADL tasks are a long way from reaching the consumer market. There are several reasons for this. First, the majority of these tasks remain challenging for today's robots, as they require a high degree of manual dexterity, sensing capability, prior task knowledge, and learning capability. Furthermore, most autonomous, proximate robots move extremely slowly due to safety and computational purposes, which will undoubtedly be frustrating for end-users. Finally, even if robots could perform some of these more complex ADL tasks, their power budgets may make them impractical for deployment in most care settings.

However, there have been substantial gains in recent years for other tasks. For example, robots that provide DRUs with additional physical reach (e.g., smart on-body prostheses, wheelchair-mounted robot arms) and robots which provide multi-setting mobility capability (e.g., exoskeletons, accessible personal transportation devices) \cite{okamura}. These are likely to continue to be the types of systems which reach end-users first for the foreseeable future.

\subsection{Cognitive Tasks}
\subsubsection{Clinicians}
\label{training}

Any technology that can effectively reduce clinical workload is likely to be warmly-embraced in healthcare. Many of these systems exist in a non-embodied fashion, for example, decision support tools to aid in emergency medicine \cite{gonzales2015designing}, patient logistical management, or charting. However, robotic systems may have a place within this domain, particularly if a robot is well-integrated into existing workflow and able to access EHR data. For example, perhaps a medication management robot could anticipate a clinician's ``next move'' in treatment by pre-fetching a likely medicine from the pharmacy. Or perhaps a robot could deliver personalized messages to family members in waiting rooms to update them on the status of their relative while clinicians are occupied with other tasks. 

Another area where robotics has been extensively used to aid clinicians with cognitive tasks is in clinical simulation and training. Robotic patient simulators are life-sized, humanoid robots which can breathe, bleed, speak, expel fluids, and respond to medications. They are the most commonly used humanoid robot worldwide, and provide learners with the ability to simultaneously practice both procedural and communication skills \cite{iva2014}. These robots are used by inter-professional clinicians across a wide range of specialties, including acute care, perioperative care, trauma, and mental health care. The author and her students have been designing the next generation of these simulators, which can convey realistic facial patient pathologies, such as pain, stroke, and cerebral palsy, and are integrated with on-board physiological models \cite{iva2014,riek-mental-health}.

\subsubsection{Direct Robot Users and Care Givers}

The ways in which robots may be able to provide cognitive task support to CGs has yet to be fully realized. However, similar to clinicians, the ability to reduce cognitive load would be greatly welcomed. CGs in particular are often overburdened when providing care; they frequently have other family members to care for, other jobs, and their own lives (and health) to manage \cite{bastawrous2013caregiver}. Robots might be able to cognitively support CGs by learning and anticipating their needs, prefetching items, attending to time-intensive tasks which detract from care, etc.

For DRUs, robotics technology might be able to help facilitate independence by providing sensory augmentation or substitution. For example, DRUs who are blind or low vision may benefit from a robotic wayfinding tool, or DRUs using robotic prostheses might receive sensory feedback from a robotic finger in their shoulder. 

Robots also may be able to help DRUs with regaining (or supplementing) cognitive function in neurorehabilitative settings, such as in cases of stroke, post-traumatic stress disorder, or traumatic brain injury. Robots also may provide socio-emotional support to DRUs: to provide companionship, teach people with autism to learn to read emotions, or to help reduce symptoms of dementia. However, there is a paucity of clinical effectiveness trials showing DRU benefit compared to standard treatment, so it is unclear what the future for these robots may be \cite{riek-mental-health}.  

\begin{figure*}[t]
\centering
\includegraphics[width=\textwidth]{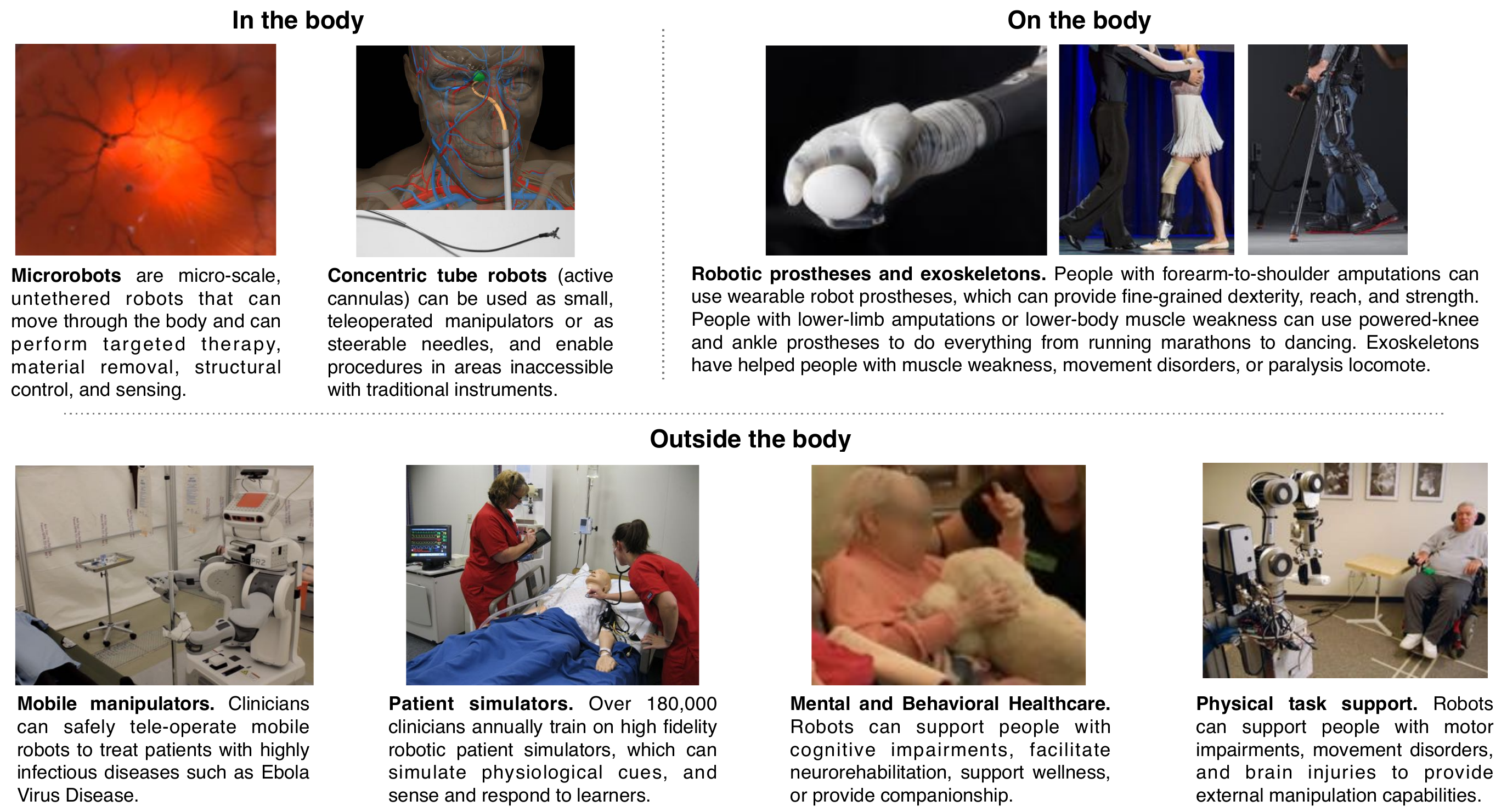}
\caption{Key examples of recent adcances in healthcare robotics. Those inside and on the body are primarily intended for direct robot users, and those outside the body for direct robot users, care givers, and clinicians. These robots have the potential to be used across a range of care settings and clinical foci, and can provide both physical and cognitive support. \textit{Image credits (clockwise from upper left): B. Nelson, R. Alterovitz, Mobius, TED, Ekso Bionics, B. Smart, L. Riek, S. Sabanovic, C. Kemp.}}
\label{fig:examples}
\end{figure*}

\section{Recent Advances in Healthcare Robotics}
\label{section3}

The 2016 US Robotics Roadmap was recently released \cite{roadmap2016}, which frames the state of the art in robotics and future research directions in the field. Over 150 robotics researchers contributed, including the author of this article. The roadmap includes a detailed summary of advancements in robotics relating to health and wellness. Some key focus areas include: aging and quality of life improvement, surgical and interventional robotics, rehabilitative robotics, and clinical workforce support.

In general, robots used in these areas can be divided into three categories: inside the body, on the body, and outside the body. Those inside and on the body are primarily intended for direct robot users, and those outside the body for direct robot users, care givers, and clinicians.  These robots have the potential to be used across a range of care settings and clinical foci, and can provide both physical and cognitive support.

\subsection{Inside the body}

Recent advances for internal robots have occured in the fields of microrobotics, surgical robotics, and interventional robotics. Microrobotics are micro-scale, untethered devices which can move through the body and can perform a range of functions, such as targeted therapy (i.e., localized delivery of medicine or energy), material removal (e.g., biopsy, ablation), structural control (e.g., stent placement), and sensing (e.g., determining oxygen concentrations, sensing the presence of cancer) \cite{micro}. Recent advances in the field have enabled actuating, powering, and controlling these robots (see Nelson et al. \cite{micro} for a  review.)

In surgical and interventional robotics, a range of advances have been made that enable clinicians to have improved dexterity and visualization inside the body and reduce the degree of movement during operations   \cite{roadmap2016}. Furthermore, promising advances have been made in concentric tube (active cannula) robots. These robots are comprised of pre-curved, concentrically nested tubes which can bend and twist throughout the body. The robots can be used as small, teleoperated manipulators or as steerable needles. The robots can enter the body directly, such as through the skin or via a body opening, or could be used via an endoscope \cite{gilbert2016concentric}.

Some future research directions for in-the-body robots include new means for intuitive physical and cognitive interaction between the user and robot, new methods for managing uncertainty, and providing 3D registration in real-time while traversing both deformable and non-deformable tissue
\cite{roadmap2016}

\subsection{On the body}

In terms of wearable robots for DRUs, there have been recent advances in the areas of actuated robot prostheses, orthoses, and exoskeletons. A \textit{prothesis} supplants a person's missing limb, and acts in series with a residual limb. An \textit{orthosis} is a device that helps someone who has an intact limb but an impairment, and an \textit{exoskeleton} provides either a person with intact limbs (DRU or otherwise) assistance or enhancement of existing physical capability. Orthoses and exoskeletons act in parallel to an existing limb \cite{tucker2015}. 

All of these robots can be used to enable DRUs to perform tasks. For example, people with forearm-to-shoulder amputations can use wearable robot prostheses, which can provide dexterity, reach, and strength. People with lower-limb amputations or lower-body muscle weakness can use powered-knee and ankle prostheses to engage in activities ranging from everyday mobility to running marathons and dancing. Exoskeletons have helped people with muscle weakness, movement disorders, and paralysis locomote. 

Several advances have been made recently in how people interface with these robots. For example, some robot prostheses offer neural integration to provide tactile feedback and increasingly more intuitive control of the limb \cite{roadmap2016}. Other advances include an increase in the workspace and range of motion of wearable robots, as well as improvements in user comfort.

\subsection{Outside the body}

Robots outside the body are being used across many clinical application spaces. For clinicians,  mobile manipulators are being used to help treat patients with highly infectious diseases \cite{kraft2016}, aid in remote surgical procedures \cite{okamura}, and help provide physical assistance to CLs when moving patients \cite{mukai2010development}. They are also used extensively in clinical training, as discussed in Section \ref{training}.

Robots are also being explored in mental and behavioral healthcare applications. Robots are being used to support people with autism spectrum disorder and cognitive impairments, to encourage wellness, and to provide companionship \cite{riek-mental-health}. (See Riek \cite{riek-mental-health} for a detailed review of these applications). 

For physical task support, robots can provide external manipulation and sensing capabilities to DRUs. For example, wheelchair-mounted robot arms can provide reach, smart wheelchairs can help facilitate safe navigation and control, and telepresence robot surrogates can enable people with severe motor impairments the ability to fly, give TED talks, and make coffee \cite{tom, chen2013robots, tsui2011want}.

There are other examples of external robots that are outside the scope of this paper, but could prove highly pertinent in healthcare. For example, autonomous vehicles may provide new opportunities for DRUs to locomote, or may enable EMTs to focus on treating patients rather than driving ambulances. Telepresence may also have unforeseen applications in healthcare, such as through aerial manipulation, drone delivery of medical supplies, etc.


\section{Healthcare Robotics Adoption: Challenges and Opportunities}
\label{adopt}
While there are exciting advances in healthcare robotics, it is important to carefully consider some of the challenges inherent in healthcare robotics, and discuss ways to overcome them. Robots have the ability to enact physical change in the world, but in healthcare that world is inherently safety critical, populated by people who may be particularly vulnerable to harm due to their disability, disorder, injury, or illness. Stakeholders face five major considerations when considering deploying robots in healthcare: Usability and Acceptability, Safety and Reliability, Capability and Function, Clinical Effectiveness, and Cost Effectiveness. Each is explored below.

\subsection{Usability and Acceptability}
\label{usability}

Robots that are difficult for primary stakeholders to use have a high likelihood of being abandoned. This phenomenon has been well documented in the Assistive Technology Community
 \cite{Lu2011, dawe2006desperately, brose2010role}. For example,
a 2010 study reported that as many as 75\% of hand rehabilitation robots were never actually tested with end users, rendering them completely unusable in practice and abandoned \cite{Balasubramanian2010}.  

One the major challenges is that clinicians, even those who are well-educated and accomplished in their disciplines, often have low technology literacy levels \cite{lluch2011healthcare}. Thus, if they themselves find a robot unusable, the likelihood of them successfully training a diret robot user or care giver to use the robot is greatly diminished. 

Another challenge is that DRUs are often excluded from the robot design process, which leads to unusable and unsuitable technology. Robots with multiple degrees of freedom, such as wearable prostheses or wheelchair-mounted arms, require a high level of cognitive function to control \cite{tsui2011want}. However, many people needing such robots often have co-morbidities (i.e., other conditions), which can make control a further exhausting process. 

There are several ways to address this issue. One approach is for robot makers to reduce robot complexity. Balasubramanian et al. \cite{Balasubramanian2010} argue for functional simplicity in therapeutic robot design, which will lead to robots that are easier for all primary stakeholders to use, control, and maintain. This concept is echoed in much of the reliability and fault tolerance literature; lower-complexity robots are more likely to be longitudinally reliable and fault tolerant. 

Forlizzi and Zimmerman propose the idea of a service-centered design process, wherein rather than only think about a single user and a system, designers consider including the broader ecosystem surrounding a technology \cite{forlizzi2013promoting}. This is a particularly beneficial idea in healthcare robotics. Rarely will there be one DRU and one robot; rather, there is a complex social structure surrounding caregiving that should be considered carefully in robot design.

Another important barrier to healthcare robot adoption is its acceptability. The morphology, behavior, and functionality of a robot play a major role in its adoption and use. When a DRU uses a robot in public, they are immediately calling attention to their disability, disorder, or illness. DRUs already face significant societal stigma, so frequently avoid using anything which further advertises their differences, even if it provides a health benefit \cite{shinohara2011shadow,riek2011using,parette2004assistive}.  

Shinohara and Wobbrock argue that in addition to designers considering the functional accessibility of system, they also consider its social accessibility, and employ a ``Design for Social Acceptance'' (DSA) approach \cite{shinohara2011shadow}
This means going beyond purely functional designs, which may be ``awkward and clunky''  \cite{shinohara2012new}. Robot makers are usually primarily concerned about a robot's functional capabilities; e.g., can the robot perform its task safely and reliably given workspace, environmental, and platform constraints. However, the aforementioned literature suggests that there may be great value in also considering a robot's appearance and behavior to help enable technology adoption.

\subsection{Safety and Reliability}

When robots and people are proximately located, safety and reliability are incredibly important. This is even more critical for DRUs who may rely extensively on robots to help them accomplish physical or cognitive tasks, and who may not have the same ability to recover from robot failures as easily as non-DRUs. 

There has been a fair bit of work on safe physical human-robot interaction, particularly with regard to improving collision avoidance, passive compliance control methods, and new advances in soft robotics to facilitate gentle interaction
 \cite{trivedi2008soft}. There also have been recent advances on algorithmic verifiability for robots operating in partially unknown workspaces \cite{hadas16}, which may prove fruitful in the future.

However, there has been little work to date on safe cognitive human-robot interaction. People with cognitive disabilities and children are particularly prone to being deceived by robots \cite{riek-mental-health}. This is an important and under-explored question in the robotics community, though a few efforts have been made recently with regard to encouraging robot makers to employ value-centered design principles. For example, ensuring the appearance of the robot is well-aligned with its function (e.g., avoiding false-advertising), enabling transparency into how a robot makes decisions, and maintaining the privacy and dignity of DRUs \cite{woody, riek-howard-ethics}.

Another way to help bridge the safety gap is for robot makrers to employ in-depth testing and training regimens which enable direct robot users, care givers, and clinicians to fully explore the capabilities of a platform. This can help prevent people from either over-relying or under-relying on the robot, and help facilitate trust. 
%

\begin{figure}[t]
\centering
\includegraphics[width=\columnwidth]{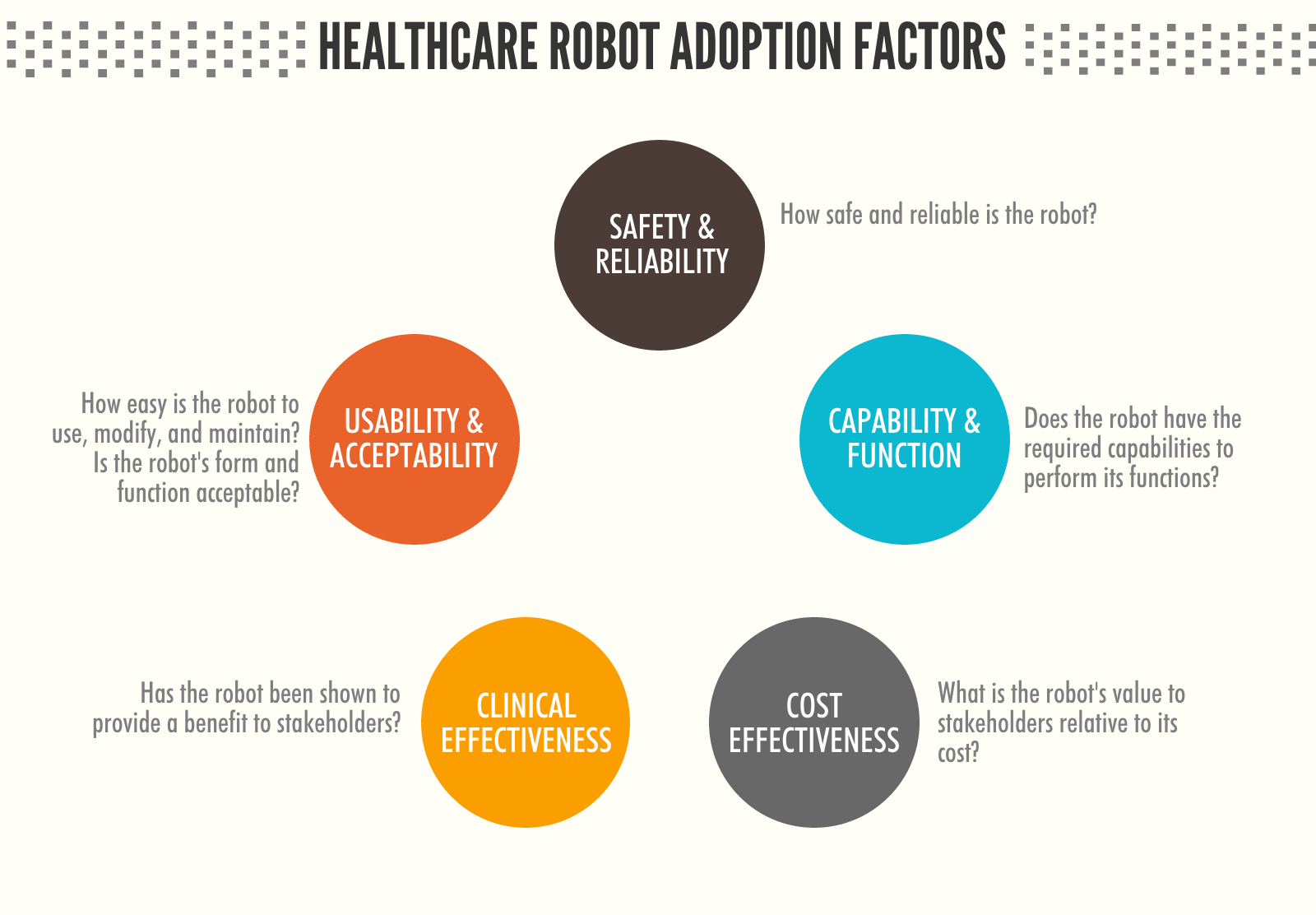}
\caption{Factors that will affect the widespread adoption of robotics in healthcare.}
\label{fig:adopt}
\end{figure}

\subsection{Capability and Function}

The field of robotics has seen amazing capability gains in recent years, some of which have been instrumental in healthcare. However, despite these advances, robotics is still an exceptionally difficult problem. For example, many demonstrations in robotics technology remain demos, and fail outside of highly constrained situations \cite{Christensen2016}. This is particularly problematic when designing technology for healthcare: most problems are open-ended, and there is no ``one-size-fits-all'' solution \cite{gonzales2015designing, riek-rss}. Every person, task, and care setting are different, and require robots to be able to robustly learn and adapt on the fly.

As discussed in Section \ref{place}, care settings differ substantially. Even the same type of care setting, such as an emergency department or assisted living facility, have substantial differences in their environment, practices, and culture. In our prior work designing health information technology, we have demonstrated that these differences can be surmounted by conducting multi-institutional trials, and by building solutions that are adaptable to different care settings  \cite{gonzales2016visual}. The same approach can be taken in robotics.

Real-world, real-time,  robust perception in human environments is a another major challenge in robotics. While the field of computer vision has seen advances in solving still-image, fixed-camera recognition problems, those same algorithms perform poorly when both the cameras and people are moving, data is lost, sensors are occluded, or there is clutter in the environment. However, these situations are highly likely in human social settings, and it is an open challenge to sense, respond to, and learn from end users in these settings \cite{riek-rss}. There have been some recent advances, however: the fields of social signal processing and human-robot interaction have moved toward multi-modal sensing approaches, which help enable more robust algorithms. Furthermore, lifelong learning and longitudinal experimental approaches have also enabled researchers to surmount some of these perceptual challenges. 
Modeling situational context and object and environmental affordances within them can also be a useful tool in surmounting these issues \cite{riek-rss, roadmap2016}.

Learning, too, is a challenge. It is critical that primary stakeholders, who have a wide range of physical abilities, cognitive abilities, and technology literacy levels, are able to easily repurpose or reprogram a robot without a RM present. This level of adaptability and accessibility presents robot makers with a complex technical and socio-technical challenge. As mentioned in Section \ref{usability}, simple is undoubtedly better; it helps constrains the problem space and lowers the complexity of the system. Another major aid will be the research community continuing to develop new datasets, evaluation metrics, and common platforms \cite{Christensen2016}; these have shown to be useful in other computing domains, so are likely to be helpful here.

\subsection{Cost Effectiveness}

When robots are being acquired in healthcare, it is important that their cost effectiveness is considered beyond the purchase, maintenance, and training costs for the system. For example, when electronic health records (EHRs) were first employed in hospitals, they were touted as a means to save clinicians and patients time. However, because EHR systems were so poorly designed, difficult to use, and poorly integrated into existing they ended up creating substantially more non-value added work. This resulted in ``unintended consequences'', including increasing costs and patient harm \cite{jones11}. It is critical these same pitfalls are avoided for robots.

The Agency for Healthcare Research and Quality (AHRQ) created a guide for reducing these unintended consequences for EHRs \cite{jones11}; the same methodological approach can be employed for robots. For example, when assessing the acquisition and deployment of a robot in a first place:

\begin{packed-enum3} 

\item \textit{Are you ready for a robot (and is a robot ready for you)?} HAs must carefully consider their institution's robot readiness. Robots may solve some problems, but may make others worse. For example, suppose a supply-fetching robot that is purchased help nurses save time. However, it has difficulties functioning at high-volume times of day due to sensor occlusion, so supply deliveries end up being delayed. This causes a cascade effect, increasing the workload of nurses. Situations like these can be remedied through a careful exploration of existing workflow in a unit, and by fully understanding a robot's existing capabilities and limitations. See \cite{gonzales2015designing,gonzales2016visual} for examples on engaging in this process with clinicians in safety critical settings.  

\item \textit{Why do you want a robot?} It is important stakeholders define exactly why a robot is necessary for a given task in the first place. What are the goals of the stakeholders? What is the plan for deploying the robot, and how will success be measured?  These questions can also be explored through design activities while assessing workflow and institutional readiness. 

\item \textit{How do you select a robot?} As mentioned previously, functionality is only one aspect to a robot; there is also: usability, acceptability, safety, reliability, and clinical effectiveness. While there are not yet definitive guidelines to aid HAs in this process, science policy is starting to be shaped within this space. The CCC recently held an event entitled ``Discovery and Innovation in Smart and Pervasive Health''
\footnote{{\url{http://cra.org/ccc/events/discovery-innovation-smart-health/}}}, which brought together over 60 researchers from across academia, industry, and government, many of whom are roboticists who work in health. These efforts will hopefully begin to provide guidelines in the future.

\item \textit{What are the recommended practices for avoiding unintended consequences of robot deployment?} Successfully deploying robots is a difficult process that may result in a disruptive care setting, and upset key stakeholders. To avoid unintended consequences, it is important that:
\begin{packed-enum3}
	\item The robot's scope is well-defined with clear goals
	\item Key stakeholders are included and engaged in the deployment from the onset
	\item Detailed deployment plans are provided but are not overly complicated.
	\item There are multiple ways to collect, analyze, and act on feedback from users
	\item Success metrics should be determined in advance and evaluated continually
	\item Quality improvement should be supported on an ongoing basis
\end{packed-enum3}

Recently, the IEEE released a document on ``Ethically Aligned Design'' which contains detailed suggestions for how to engage in this value-centered practice in engineering, which could be helpful for all stakeholders moving forward
\footnote{\url{standards.ieee.org/develop/indconn/ec/ead_v1.pdf}}


\end{packed-enum3}

\subsection{Clinical Effectiveness}
 
Clinical effectiveness answers the question ``does it work?'' In particular, does a given intervention provide benefit to a primary stakeholder? This question is answered by conducting thorough, evidence-based science. For robots directly affecting DRUs, this evidence comes from comparative effectiveness research (CER), which is ``generated from research studies that compare drugs, medical devices, tests, surgeries, or ways to deliver health care.''\footnote{Agency for Health and Research Quality, \url{effectivehealthcare.ahrq.gov/index.cfm/what-is-comparative-effectiveness-research1/}} CER can include both new clinical studies on effectiveness, or can synthesize the existing literature in a systematic review. 

All consumer in-the-body robots and many on-the-body robots must undergo regulatory approval before they can be marketed and sold. In the United States, this approval is through the FDA, which typically requires a strong level of evidence showing the effectiveness and safety of a medical device. Outside-the-body robots typically do not need to undergo a device review process provided they fall within existing classifications; for example, Paro the robot seal (see Figure \ref{fig:examples}, bottom right) is classified by the FDA a neurological therapeutic device, and thus is exempt from premarket review.  Shimshaw et al. \cite{drew} argue this lack of regulation of healthcare robots may be harmful to stakeholders both physically and informationally, and should be subject to premarket review on dimensions including privacy, safety, reliability, and usability. 

In the meanwhile, while the policy community races to catch up with technology, the robotics community can and should engage in research that tests the clinical effectiveness of  robots across care settings. Begum et al. \cite{begum2016robots} suggest robot makers follow existing clinical effectiveness benchmarks within their intended care space and adopt them for use with robots. Furthermore, Riek \cite{riek-mental-health} suggests that when conducting CER with robots, particularly in cognitive support settings, it is not sufficient to simply test robot vs. no-robot, as the morphology can affect outcomes, but to instead to test actuated vs. non-actuated.

\section{Discussion}

Healthcare robotics is an exciting, emerging area that can benefit all stakeholders across a range of settings. There have been a number of exciting advances in robotics in recent years, which point to a fruitful future. How these robots ultimately will be integrated into the lives of primary beneficiaries remains unknown, but there is no doubt that robots will be a major enabler (and disruptor) to health.

It is critical that both the research and industrial communities work together to establish a strong evidence-base for healthcare robotics. As we have learned from the large-scale deployment of EHRs, technology development and deployment cannot happen in a vacuum, or it is likely to cause grave harm to DRUs, overwhelming stress to clinicians, and astronomical unseen costs. It is wise for all stakeholders to proceed cautiously and deliberately, and consider the full context of care as much as possible. 

It is also critical that direct robot users remain directly involved in the research, development, and deployment of future robots in health and wellness across the entire lifecycle of a project, as ultimately they are the ones who will be using these robots. As discussed earlier, ignoring DRU input leads to unusable, unsuitable, and abandoned robots, which benefits no one. Secondary and Tertiary stakeholders should look to the Patient Centered Outcomes Research Institute (PCORI){\footnote{\url{http://pcori.org}}} as a highly successful model for how to engage with primary stakeholders in clinical research and development.

Finally, it is important that robot makers work with DRUs to help bridge technology literacy gaps and appropriately set expectations. Most people's experience with robotics comes from movies or media, which rarely reflects the true state of affairs. Robots are quite fallible in the real world, and will remain so for the foreseeable future; however, they still have the potential to be a remarkable game changer in health.

\section{Acknowledgements}

Some research reported in this article is based upon work supported by the National Science
Foundation under Grant Nos. IIS-1253935, SES-1457307, and the Luce Foundation. 
The author also thanks Eric Valor, Deanna Gates, Selma Sabanovic, Bill Smart, Ron Alterovitz, and Brenna Argall.

\bibliographystyle{abbrv}
\bibliography{riek-acm-2016-bib}

\end{document}